# Occlusion-aware Visual Tracker using Spatial Structural Information and Dominant Features

Rongtai Cai[1] and Peng Zhu[2]

[1]Fujian Provincial Engineering Technology Research Center of Photoelectric Sensing Application, College of Photonic and Electronic Engineering, Fujian Normal University, China

[2]Fujian Newland Computer Co Ltd., China

**Abstract:** *To overcome the problem of occlusion in visual tracking, this paper proposes an occlusion-aware tracking algorithm. The proposed algorithm divides the object into discrete image patches according to the pixel distribution of the object by means of clustering. To avoid the drifting of the tracker to false targets, the proposed algorithm extracts the dominant features, such as color histogram or histogram of oriented gradient orientation, from these image patches, and uses them as cues for tracking. To enhance the robustness of the tracker, the proposed algorithm employs an implicit spatial structure between these patches as another cue for tracking; Afterwards, the proposed algorithm incorporates these components into the particle filter framework, which results in a robust and precise tracker. Experimental results on color image sequences with different resolutions show that the proposed tracker outperforms the comparison algorithms on handling occlusion in visual tracking.*

## 1. Introduction

Despite the advances in visual tracking in these years, there are still unsolved problems, such as background clutters, occlusions, pose changes, illumination variations. For example, a tracker tends to drift to the neighborhoods of an object's ground-truth state, when the object's appearance becomes dissimilar from that of the template while occlusion occurs.

A reasonable solution to the problem is tracking only based on these image parts that are not occluded by the disruptors. Such a strategy is employed by part-based trackers [9, 10]. The part-based trackers divide the object into image patches, figure out these patches that are occluded by measuring the appearance differences between these patches and their counterparts in the template, then exclude them from the appearance model, or limit their contributions to the appearance model. By doing so, such trackers can overcome the problem of partial occlusions to some extent.

But there are some problems in the conventional part-based trackers [9, 10]. First, there are no feasible solutions for the adaptive division of a target into its parts. Second, the previously designed algorithms do not provide the strategy to facilitate visual tracking by using image features with different reliability.

To overcome such problems, this paper proposes an occlusion-aware visual tracking algorithm, which employs an appearance model that is robust against occlusions. The proposed appearance model is composed of the dominant features of the parts and the spatial constraints among parts.

The novelties of the proposed algorithm are as follows.

First, unlike the baseline part-based tracking schemes [9, 10], the proposed algorithm divides the object into discrete parts adaptively by means of color clustering.

Second, the proposed algorithm is different from the baseline multi-cue or multi-feature tracking schemes [8, 16, 19] in that it uses a dominant feature for tracking, which is the most discriminative one selected from the features.

At last, beyond what is stated above, the benefit of the proposed scheme is beyond robust against occlusion. The proposed scheme utilizes spatial structural constraints among the image patches of the object as cues for tracking. As complementation of the appearance, the spatial structural constraint plays an important role in the proposed scheme. For example, there are false targets that are similar to the ground-truth target in the distribution of the appearance, such as Color Histogram (HoC) [7] or Histogram of Gradient Orientations (HoG) [13]. But they are different from the ground-truth target in the spatial arrangements of image patches. Such kind of false targets can be distinguished from the ground-truth target according to the spatial structural constraints among the image patches.

## 2. Related Work

Visual tracking has been an important topic in computer vision for a long time, therefore a handful of algorithms are proposed in the past decades. There are famous examples, such as mean-shift tracker [2], covariance tracker [21], color name tracker [3], Particle Filter (PF) [1, 14], kernel correlation tracker [6], and tracker based on deep neural networks [26].

These trackers must exploit features from images for tracking, either explicitly or implicitly. There are color features, edge features, textural information, motion information, and so on. But single feature for tracking tends to fail, it is likely to be confounded with false targets or background clutters. Hence, multi-feature/multi-cue trackers are proposed. For example, Lan *et al.* [8] used a joint sparse representation and robust feature-level fusion for multi-cue visual tracking. The tracker dynamically removed unreliable features to be fused for tracking and performed feature fusion on kernel spaces. Meanwhile, Sadeghian *et al.* [16] proposed a scheme to track multiple cues with long-term dependencies. They proposed a structure of recurrent neural networks jointly reasons on multiple cues. Their algorithm tracked multiple targets robustly by using targets' appearance, motion, and interactions. Dhassi and Aarab [4] proposed a visual tracker fusing multiples cues. They fused the color and texture features to describe the appearance of the object under the particle filter framework. Walia and Kapoor [19] proposed an online object tracking algorithm via an adaptive multi-cue based particle filter framework. They estimated partial conflicting masses and conjunctive consensus among three cues for each evaluated particle. They used the reliabilities of context-sensitive transductive cues for discounting the particle likelihoods for quick adaptation of tracker.

These multi-cue tracking schemes have largely improved the robustness of visual tracking, but they won't work well while there are occlusions. Hence part-based trackers are proposed. For example, Shu [18] proposed a part-based multiple-person tracker with partial occlusion handling. The tracker dynamically handled occlusions by distributing the score of the learned person classifier among its corresponding parts. By doing so, the tracker detected and predicted partial occlusions, and prevented the performance of the classifiers from being degraded. Meanwhile, Wang *et al.* [20] proposed a robust occlusion-aware part-based visual tracking with object scale adaptation. They used correlation filters to integrate the global model and part-based model. They used both global and local information to improve the robustness of the tracker. Yao *et al.* [23] proposed a part-based robust tracker using online latent structured learning. They modeled the parts of objects as latent variables and extended an online algorithm to the structured prediction case with the latent part variables.

Zhang *et al.* [24] proposed a part-based visual tracker with spatially regularized correlation filters. They used multiple correlation filters to extract features within the range of the object, alleviated the boundary effect problem, and avoided penalization of the target content.

As stated above, the part-based schemes are usually embedded into a tracker such as a particle filter or correlation filter, where particle filter is a popular visual tracking algorithm in these years. For example, Morales *et al.* [12] proposed a combined voxel and particle filter-based approach for fast obstacle detection and tracking. They used a particle filter to extract the 3-D object and estimated the motion. Meanwhile, Zhang *et al.* [25] proposed a correlation particle filter for visual tracking. They exploited and complemented the strengths of both the correlation filter and the particle filter. The tracker effectively handled large-scale variations via a particle sampling strategy, maintained multiple modes in the posterior density using little particles, and enabled the particles to stand close to the modes of a state by using a mixture of correlation filters. Fang *et al.* [5] proposed an on-road vehicle tracker based on a part-based particle filter. They combined the part-based strategy into a particle filter. They represented the central position of the vehicle as a hidden state and updated the particles corresponding to vehicle parts sharing the same motion effectively. Meng and Zhang [11] proposed an object tracker using a particle filter based on adaptive patches combining multi-features. They used weighted Bhattacharyya coefficients to calculate the sub-patch matching degrees of the particles, and adjusted the particle sub-patch weights by integrating the particle space information.

## 3. Spatial Structural Cues and Dominant Features for Tracking

### 3.1. Spatial Structural Information for Tracking

The proposed algorithm divides the object into discrete patches according to the object's pixel distribution. To simplify the division, the target image is mapped from 2D space onto 1D space. Suppose the size of the patch is $N \times M$, the mapping of a 2D patch onto 1D space is the accumulation of $N$ elements in the vertical direction:

$$C_j = \frac{1}{N} \sum_{i=1}^{N} b(i, j), \quad j = 1, 2, \dots M \,, \tag{1}$$

where $b(i,j)$ is the color value of a pixel on $(i,j)$, $C_j$ is the accumulation of $b(i,j)$ upon $i$.

The proposed algorithm then divides the object into discrete patches by clustering on columns $\{C_1, C_2, \dots, C_M\}$. If $|C_j - C_{j+1}| \leq T_c$, then $C_{j+1}$ follows into the same cluster of $C_j$. Otherwise, a new cluster is initialized by $C_{j+1}$, where $T_c$ is a preset threshold.

Another clustering and segmentation take place upon rows $\{R_1, R_2, \ldots, R_N\}$ in the horizontal direction, where $R_i$ is the accumulation of $b(i,j)$ upon $j$. An example of the division of a pedestrian is presented in Figure 1.

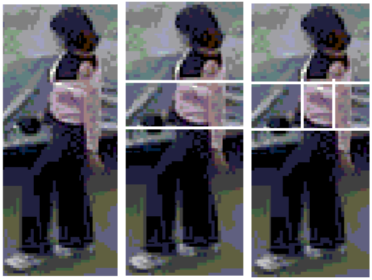

Figure 1. Segmentation of an object into discrete patches.

In Figure 1, the proposed algorithm maps the pedestrian image from 2D space onto 1D space in the vertical direction, then divides it into three patches according to the object's pixel distribution in the 1D space. These patches correspond to the head, the body, and the legs of the pedestrian respectively. Moreover, the proposed algorithm can divide a patch into sub-parts. In the case of the pedestrian, there are lots of background clutters both at the right side and the left side in the body part. The proposed algorithm can divide this patch into three sub-parts, and then exclude the left sub-part and the right sub-part from the appearance model.

The spatial arrangement of the parts of an object are relatively stationary, hence it is ideal to serve as a cue for tracking. The proposed algorithm uses a graph to model the target by representing the image parts as vertices and the spatial connections among the parts as edges. The spatial structural constraints of two parts (vertices) can be model as:

$$p(n_i, n_j) = \exp\left(-K\left(\left|n_i - n_j\right| - \xi_{ij}\right)^2\right), \tag{2}$$

Where $K$ is a parameter for model adjustment, $n_i$ is the spatial location of the $i$-th part (vertex), $\xi_{ij}$ is the initial spatial constraint of the $i$-th part and the $j$-th part generated from the template, which will be updated during the tracking process.

The proposed algorithm examines $p(n_i, n_j)$ for all pairs of parts in an object candidate to measure the likelihood according to the spatial structural constraints.

## 3.2. Dominant Feature for Tracking

In a typical multi-feature tracking procedure [8, 16], several features are extracted and fused for tracking, which produces a robust result. A discriminative feature will generate a state that is very close to the ground-truth state of the target. But this is not the case of indistinguishable features. After fusing, the estimated state may be far away from the ground-truth state of the target due to the invalid features. Hence, feature fusing enhances the robustness of tracking but reduces the tracking precision.

Unlike the baseline multi-cue/multi-feature tracking procedure [4, 19], the proposed algorithm only uses the dominant feature of a part as the cue for tracking, which aims to overcome the problem that the fusing of multiple features reduces the tracking precision.

The dominant feature is the most discriminating one in the features, which is produced by independently comparing the part with the other parts and the background patches according to the features. The dominant feature $f_b^{dom}$ of an image part (block) $b$ is obtained by:

$$f_b^{dom} = \arg\ \max_i \left\{d\left(f_b^i, f_k^i\right)\right\}, \tag{3}$$

For any image part $k$ other than the image part $b$, where $d(,)$ is the difference operator, $f_b^i$ is the $i$-th feature of the image block $b$.

The tracker locates the target according to the likelihood of the candidates, which is in turn largely dependent on the likelihood of its parts. The likelihood of a part $b$ is defined as the similarity of the part $b$ with its counterpart $\bar{b}$ in the template, which is measured by the Bhattacharyya coefficient

$$\rho\left(b, \bar{b}\right) = \left(1 - \sum_{i=1}^{N} \sqrt{f_{dom}^b(i), f_{dom}^{\bar{b}}(i)}\right)^{\frac{1}{2}}, \tag{4}$$

Where $N$ is the size of the dominant feature.

There are many handcrafted features: HoC, Scale-Invariant Feature Transform (SIFT), HOG, Local Binary Patterns (LBP), and so on, that can be produced from an image part. The proposed algorithm chooses HoC [7] and HoG [13] as the candidates for the local dominant feature for their successfulness in visual tracking in the past decades.

## 3.3. The Candidate's Likelihood

The candidate's probability of likelihood is composed of the parts' likelihoods and their corresponding weights:

$$P_{candidate} = K \exp\left(-\frac{\sum_{b=1}^{B} w_b \cdot \rho\left(f_{dom}^b, f_{dom}^{\bar{b}}\right)}{2\sigma^2}\right), \tag{5}$$

Where $w_b$ is the weight of the image part $b$, and $\sum_{b=1}^{B} w_b = 1$, supposed that the candidate is separated into $B$ image blocks. $K$ is a normalization parameter, $\sigma$ is the variance of the candidate's likelihood. The parts' likelihood is computed according to (4). Meanwhile, the parts' weights are initialized at first and updated during the tracking process according to their reliability.

Only those that most of whose unobstructed patches satisfy the similarities of appearance and spatial structural constraints serve as target candidates. By doing so, the proposed algorithm uses the spatial structure among the patches as a cue for tracking. Note that there are false targets that have the same probability density function as the ground-truth target, but have different spatial arrangements of parts. Such false targets can be excluded from the candidates by using this cue.

### 3.4. Occlusion Handling and Template Updating

The proposed algorithm can adjust the weights of the patches according to how they are occluded. Specifically, the proposed algorithm updates the parts' weights at time $t$ by

$$w_b(t) = (1-\alpha) \cdot w_b(t-1) + \alpha \cdot \Delta w_b(t), \quad (6)$$

Where $\alpha$ is the parameter used to adjust the speed of weights' updating.

If $\alpha$ is set to one, the weight of image part $b$ is only dependent on $\Delta w_b$. While, $\Delta w_b$ is only dependent on the likelihood of the part according to

$$\Delta w_b = M \cdot \exp\left(-\frac{\rho\left(f_{dom}^b, f_{dom}^{\bar{b}}\right)}{2\sigma^2}\right), \quad (7)$$

Where $M$ is the normalization parameter, and $M = \sum_{b=1}^{B} \Delta w_b$. The likelihood/similarity is dependent on the changes of the image parts' appearance, which in turn depend on the occlusions, clutters, pose changes, illumination variations, and so on. Here, occlusion is the most influential one in these interference terms. Hence, the patches contribute to the likelihood of the target reasonably according to how they are occluded.

If the change of a patch's appearance is small enough, there is no occlusion. And this patch is used to replace its counterpart in the template. If the change of a patch's appearance is large enough, it is occluded. And the counterpart in the template is kept unchanged. Otherwise, the counterpart in the template is updated by using a weighted sum of the patch in the current frame and the counterpart in the template.

### 3.5. The Proposed Visual Tracker

The spatial structural cues and the local dominant features are integrated into a particle filter framework to produce a robust visual object tracker, which is occlusion-aware and robust against appearance changes. The Particle filter uses a predict-update cycle for state estimation. It predicts the state by first producing particles via sampling from the a priori of the state, and then transfers these particles through the system state formula. Afterward, current observations are used to correct the prediction. Thus, the a posteriori probabilities of these particles are produced. The weighted sum of the particles' states is ultimately the estimated state. The implementation of the particle filter with spatial structural constraints and local dominant features results in the proposed algorithm, as shown in Algorithm 1.

*Algorithm 1: Occlusion-aware visual tracking algorithm with spatial structural constraints and dominant features.*

*1 Initialization in the first frame:*
*1.1 Divide the object into several discrete patches according to (1);*
*1.2 Initial the state of the target and the states of the patches.*
*2 From the second frame to the last frame, do the followings:*
*2.1 Prediction: produce several particles (object candidates);*
*2.2 Measure the similarity of an image patch between every particle and the template;*
*(a)Divide the object into several patches according to (1);*
*(b)Measure every patches' likelihood using dominant features by (4);*
*(c)Measure the patches' spatial consistencies according to (2);*
*(d)Measure the likelihood between the particle and the template*
*based on (b), (c) and (5);*
*2.3 Adjust the weights of particles according to their likelihoods;*
*2.4 Estimate the final state of the object using the weighted sum of the particles' states;*
*2.5 Resample if it's necessary according to the particles' diversity;*
*2.6 Update the template of the object.*

## 4. Experimental Results

To verify the performance of the proposed tracker. First, the differences between the proposed tracker and traditional particle filter using HoC and HoG are studied, where the "Bolt2" sequence is used in this experiment to illuminate the improvement of the proposed tracker. There are 293 color images in the "Bolt2" sequence with resolutions of 480×270. There are similar false targets and heavy occlusions in the sequence. The experimental results are shown in Figure 2.

Traditional particle filter uses HoC or HoG of the whole object as a cue for tracking, which involves a lot of background clutters and is unable to discriminate false targets similar to the ground-truth target on HoC or HoG. However, while the HoC or HoG of a false target is more similar to the template than the ground-truth target does due to the target's pose changes, the tracker will mistake the false target as the ground-truth target, see the 35[th] frame in Figure 2 for example. But the proposed tracker is different. The proposed tracker excludes the clutter patches in the bounding box from the appearance model, or limit their contributions to the appearance model. And the proposed tracker updates the templates of the patches according to the object's pose changes. Hence, the proposed tracker can find the ground-truth target even under very poor conditions.

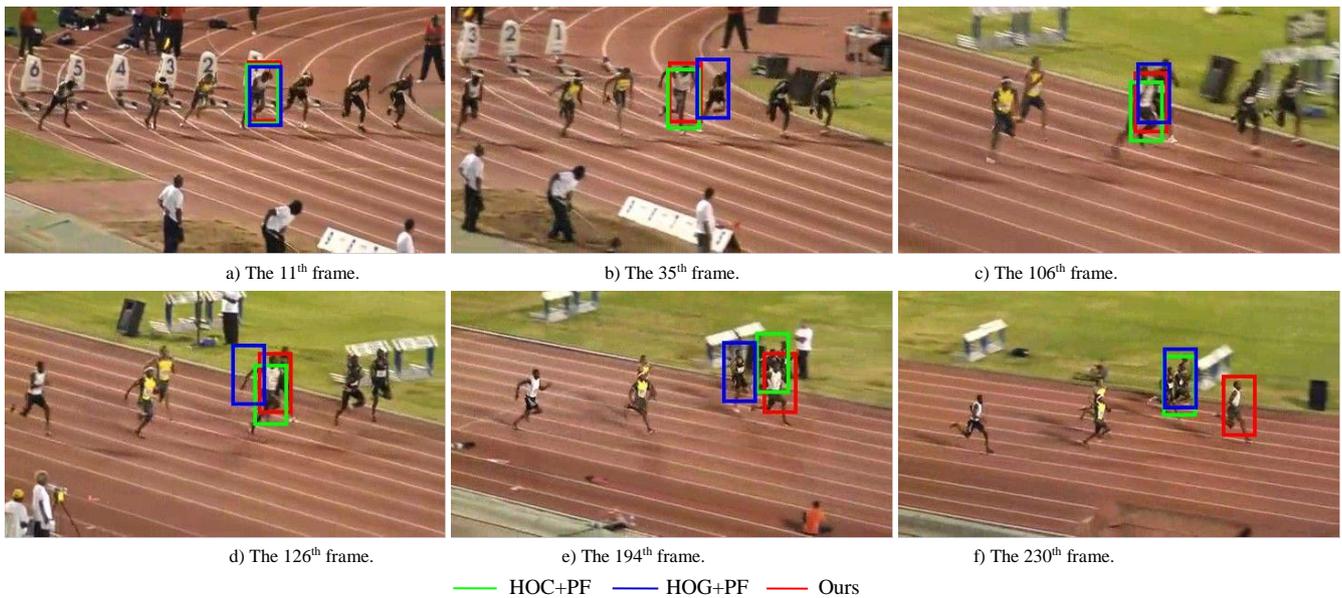

a) The 11th frame.          b) The 35th frame.          c) The 106th frame.

d) The 126th frame.          e) The 194th frame.          f) The 230th frame.

— HOC+PF   — HOG+PF   — Ours

Figure 2. Tracking results of the "Bolt2" sequence.

Furthermore, the performance of the proposed tracker is evaluated by testing it as well as three comparison tracking algorithms: Robust Superpixel tracker (RS) [22], Distribution Fields for Tracking (DFT) [17], and PF [15] on more challenging image sequences "David3" and "Woman". There are 250 color images in the "David3" sequence, where there are close background interference and, large pose changes of target. While there are 300 color images in the "Woman" sequence, and there are partial occlusions and large pose changes of the target in this sequence as well. The size of the images in the "David3" sequence is $640 \times 480$. And the size of the images in the "Woman" sequence is $384 \times 288$. The experimental results are given in Figures 3 and 4, respectively.

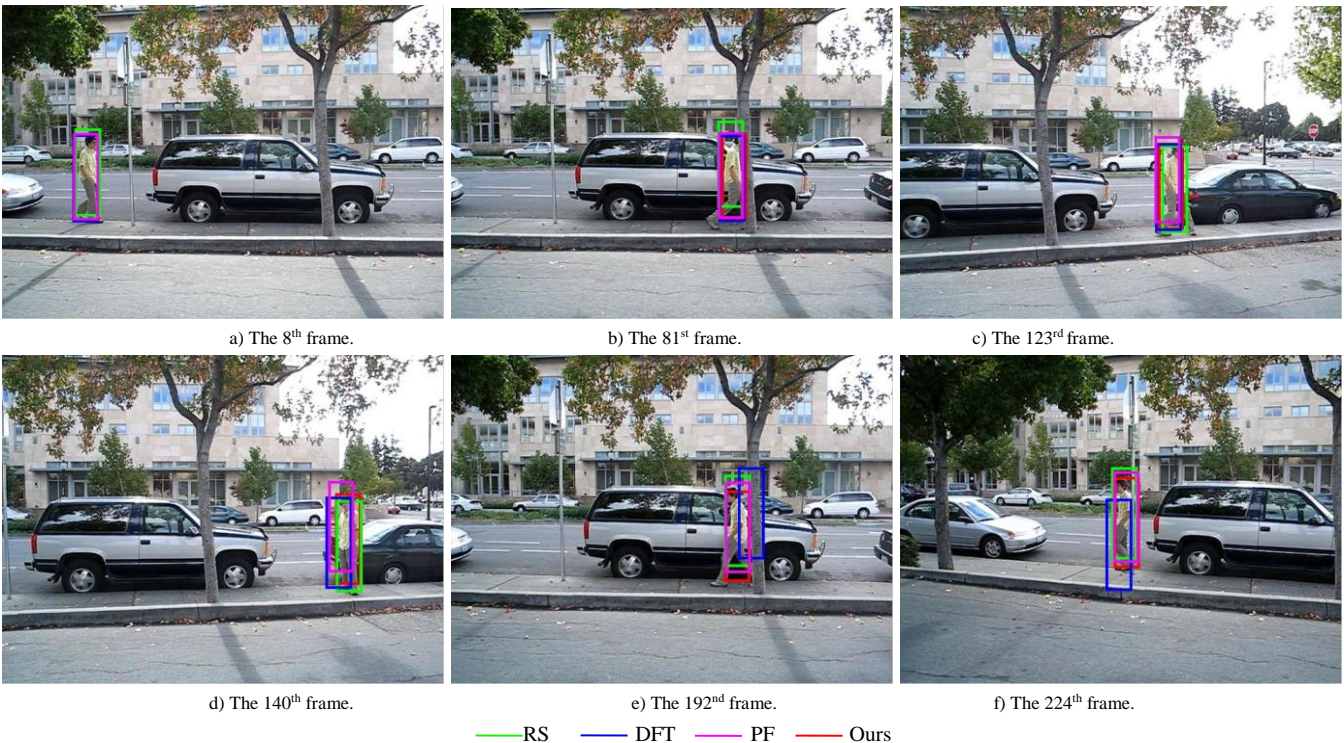

a) The 8th frame.          b) The 81st frame.          c) The 123rd frame.

d) The 140th frame.          e) The 192nd frame.          f) The 224th frame.

— RS   — DFT   — PF   — Ours

Figure 3. Tracking results of the "David3" sequence.

The experimental results of the "David3" sequence are shown in Figure 3. All of the four trackers employed in the experiments do well at the beginning. But, while the object is occluded by a tree, the RS tracker drifts away from the ground-truth, see the 81st frame for example. While the object is confused with background clutter from the rear of a black car, the PF drifts away from the ground-truth, see the 140th frame for example. Similarly, the DFT tracker is sensitive to the interference both from the tree's occlusion and the background clutters, see the 192nd frame and the 244th frame for example. But the proposed tracker shows great robustness against the interference in this experiment.

The experimental results of "Woman" produced by the four trackers are presented in Figure 4. Here, the PF tracker is sensitive to the occlusion from the white car and drifts away from the ground-truth in the 108th frame. The RS tracker is disturbed by the occlusion from the white car and drifts away from the ground-truth after long time occlusion. But the proposed tracker and DFT tracker are robust against the interference.

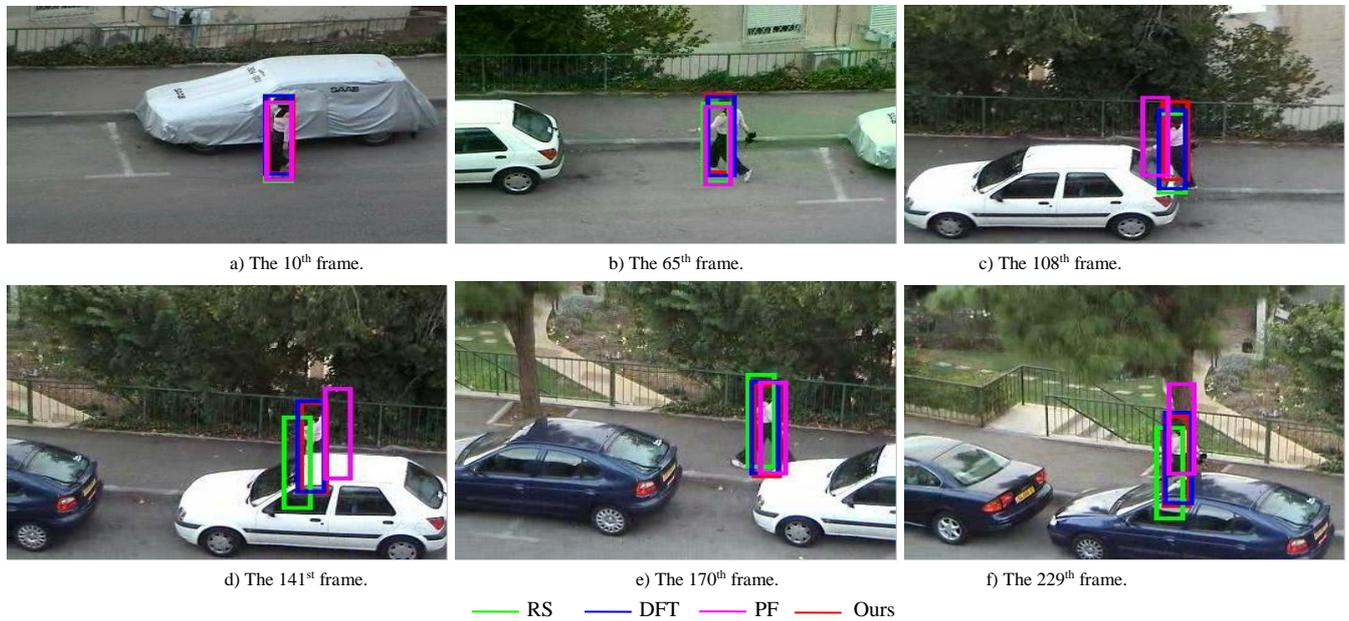

a) The 10th frame.          b) The 65th frame.          c) The 108th frame.

d) The 141st frame.         e) The 170th frame.         f) The 229th frame.

——— RS      ——— DFT      ——— PF      ——— Ours

Figure 4. Tracking results of the "Woman" sequence.

Moreover, the Center Location Error (CLE) is employed in the experiment for quantitative evaluation:

$$e(t) = \sqrt{\left(x(t) - x_g(t)\right)^2 + \left(y(t) - y_g(t)\right)^2} \ , \qquad (8)$$

Where $(x(t), y(t))$ is the estimated center of the object at frame $t$, $(x_g(t), y_g(t))$ is the ground-truth center of the object at frame $t$, $e(t)$ is the center location error. The location errors of the "David3" sequence and the "Woman" sequence are given in Figure 5.

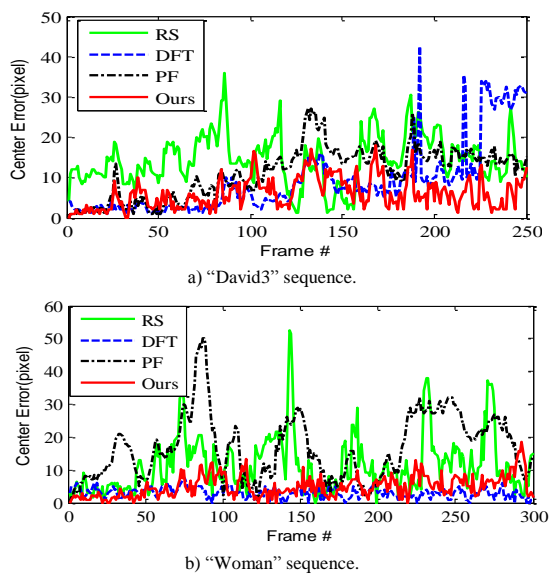

a) "David3" sequence.

b) "Woman" sequence.

Figure 5. Location errors of the "David3" sequence and the "Woman" sequence.

For the "David3" sequence, the proposed tracker and the DFT tracker outperform the RS tracker and the PF tracker at the beginning. As time goes by, the DFT tracker degrades with a large margin due to the disturbing from occlusions and background clutters. But it is not the case for the proposed tracker. Hence, the proposed tracker outperforms all of the comparison trackers in the "David3" sequence. See Figure 5-a for details.

The location errors of the "Woman" sequence are given in Figure 5-b. In this experiment, the proposed tracker and the DFT tracker outperform the RS tracker and the PF tracker. The DFT tracker is as robust as the proposed tracker. Though, the DFT tracker outperforms the proposed tracker in precision in this sequence. But considering the performance of the DFT tracker in the "David3" sequence, it is clear that the proposed tracker is more robust than the DFT tracker.

The quantitative results on the "David3" sequence are summarized in Table 1. Here, a successful ratio as the ratio of the number of frames with errors less than 16 pixels to the total frames is defined and applied to evaluate the performance of the trackers.

Table 1. Quantitative comparison of the "David3" sequence.

| Tracker | Successful Frames | Successful Ratio | Average Tracking Error |
|---------|-------------------|------------------|------------------------|
| RS | 156 | 62.4% | 10.7583 |
| DFT | 219 | 87.6% | 5.6472 |
| PF | 196 | 78.4% | 9.0366 |
| Our | 240 | 96% | 6.7711 |

At the same time, the quantitative results on the "Woman" sequence are summarized in Table 2.

Table 2. Quantitative comparison of the "Woman" sequence.

| Tracker | Successful Frames | Successful Ratio | Average Tracking Error |
|---------|-------------------|------------------|------------------------|
| RS | 251 | 71.67% | 7.2088 |
| DFT | 300 | 100% | 3.4848 |
| PF | 139 | 46.33% | 8.3088 |
| Our | 289 | 96.33% | 8.2088 |

The results in Tables 1 and 2 agree with what is stated above. According to the results, it is evident that the proposed tracker outperforms the DFT tracker, the RS tracker, and the PF tracker on robustness.

## 5. Conclusions

This paper has developed an occlusion-aware tracking algorithm. Experiments on several image sequences prove that the proposed tracker outperforms the comparison algorithms on robustness. The improvement of the proposed algorithm over the comparison algorithms lies in the employments of the dominant features and the part-based tracking strategy.

Conclusively, there are several advantages of the proposed tracker. First, the proposed tracker processes the robustness as PF due to the employment of the PF framework. Second, the proposed tracker is robust against occlusions due to the employment of the strategy of part-based tracking. Third, the proposed tracker is more precise than similar tracking algorithms due to only the dominant features are used for tracking. At last, the proposed tracker is more guaranteed to hit the ground-truth object due to the employment of the spatial structural constraints among the image patches of the object. These advantages are also verified by the experiments. The authors expect future work to exploit the power of the proposed tracker more deeply and explain more explicitly how the particle filter framework, the strategy of part-based tracking, and the dominant features work together to track the object correctly and robustly.

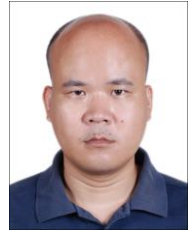

**Rongtai Cai** received the B.Eng. degree from Jilin University, Changchun, China, in 2003, and the Ph.D. degree from the Graduate University of Chinese Academy of Sciences, Beijing, China, in 2008. He was a visiting scholar with the University of Oxford, Oxfordshire, UK, from 2014 to 2015. He is currently an associate professor with the College of Photonic and Electronic Engineering, Fujian Normal University, Fuzhou, China. His research interests include computer vision and machine learning, cortex-like visual computation, and the neural computation mechanisms of vision.

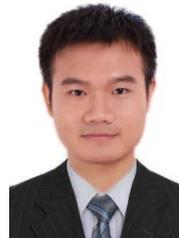

**Peng Zhu** received his engineer diploma degree in electronic and information engineering and MSc diploma degrees in Digital image processing, both from Fujian Normal University, in 2013, and 2016, respectively. Then, he joined Fujian Newland Computer Co Ltd. as a research engineer. His main fields of interest are computer vision and machine learning.